\documentclass[11pt,a4paper]{article}
\usepackage[hyperref]{acl2019}
\usepackage{times}
\usepackage{latexsym}
\usepackage{amssymb}
\usepackage{latexsym}
\usepackage{natbib}
\usepackage{makecell}
\usepackage{svg}
\usepackage{subfig}
\usepackage{multirow}
\usepackage{cleveref}

\usepackage{url}

\usepackage{float}
\usepackage{placeins}

\aclfinalcopy 

\title{Gendered Ambiguous Pronouns Shared Task: Boosting Model Confidence by Evidence Pooling}

\author{Sandeep Attree \\
  New York, NY \\
  \texttt{sandeep.attree@gmail.com} \\}

\date{}

\begin{document}
\maketitle
\begin{abstract}
  This paper presents a strong set of results for resolving gendered ambiguous pronouns on the Gendered Ambiguous Pronouns shared task. The model presented here draws upon the strengths of state-of-the-art language and coreference resolution models, and introduces a novel evidence-based deep learning architecture. Injecting evidence from the coreference models compliments the base architecture, and analysis shows that the model is not hindered by their weaknesses, specifically gender bias. The modularity and simplicity of the architecture make it very easy to extend for further improvement and applicable to other NLP problems. Evaluation on GAP test data results in a state-of-the-art performance at 92.5\% F1 (gender bias of 0.97), edging closer to the human performance of 96.6\%. The end-to-end solution\footnote{The code is available at \url{https://github.com/sattree/gap}} presented here placed 1st in the Kaggle competition, winning by a significant lead.
\end{abstract}

\section{Introduction}
The Gendered Ambiguous Pronouns (GAP) shared task aims to mitigate bias observed in the performance of coreference resolution systems when dealing with gendered pronouns. State-of-the-art coreference models suffer from a systematic bias in resolving masculine entities more confidently compared to feminine entities. To this end, \citeauthor{webster2018mind} \shortcite{webster2018mind} published a new  GAP dataset\footnote{\url{https://github.com/google-research-datasets/gap-coreference}} to encourage research into building models and systems that are robust to gender bias.

\begin{figure}[htbp]
  \centering
  \includegraphics[width=100pt]{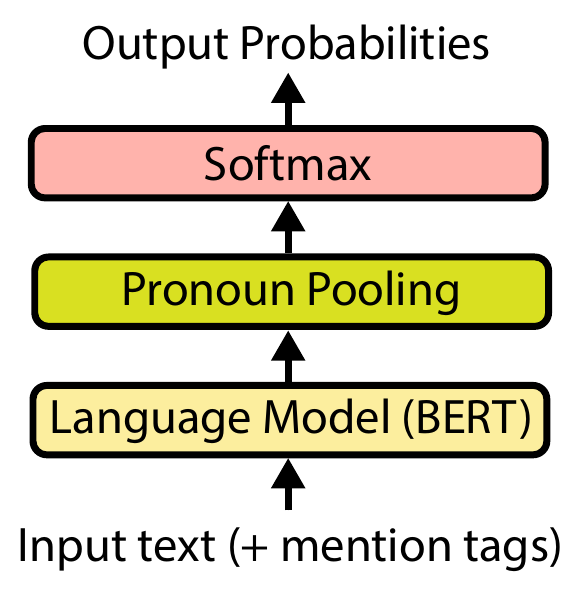}
  \caption{\label{probert}ProBERT: Pronoun BERT. Token embeddings corresponding to the labeled pronoun in the input text are extracted from the last layer of the language model (BERT) and used for prediction.}
\end{figure}

The arrival of modern language models like ELMo \cite{peters2018deep}, BERT \cite{devlin2018bert}, and GPT \cite{radford2018improving}, have significantly advanced the state-of-the art in a wide range of NLP problems. All of them have a common theme in that a generative language model is pretrained on a large amount of data, and is subsequently fine-tuned on the target task data. This approach of transfer learning has been very successful. The current work applies the same philosophy and uses BERT as the base model to encode low-level features, followed by a task-specific module that is trained from scratch (fine-tuning BERT in the process).


GAP shared task presents the general GAP problem in \textit{gold-two-mention} \cite{webster2018mind} format and formulates it as a classification problem, where the model must resolve a given pronoun to either of the two given candidates or neither\footnote{There is a case in the GAP problem where the pronoun in question may not be coreferent with either of the two mentioned candidates. Such instances will be referred to as \textit{Neither}.}. \textit{Neither} instances are particularly difficult to resolve since they require understanding a wider context and perhaps a knowledge of the world. A parallel for this case can be drawn from Question-Answering systems where identifying unanswerable questions confidently remains an active research area. Recent work shows that it is possible to determine lack of evidence with greater confidence by explicitly modeling for it. Works of \citet{zhong2019coarse} and \citet{kundu2018nil} demonstrate model designs with specialized deep learning architectures that encode evidence in the input and show significant improvement in identifying unanswerable questions. This paper first introduces a baseline that is based on a language model. Then, a novel architecture for pooling evidence from off-the-shelf coreference models is presented, that further boosts the confidence of the base classifier and specifically helps in resolving \textit{Neither} instances. The main contributions of this paper are:
\begin{itemize}
\item Demonstrate the effectiveness of pretrained language models and their transferability to establish a strong baseline (ProBERT) for the \textit{gold-two-mention} shared task. 
\item Introduce an Evidence Pooling based neural architecture (GREP) to draw upon the strengths of off-the-shelf coreference systems.
\item Present the model results that placed 1st in the GAP shared task Kaggle competition.
\end{itemize}

\section{Data and Preprocessing}
Table \ref{table:corpus_stats} shows the data distribution. All datasets are approximately gender balanced, other than stage 2 test set. The datasets, gap-development, gap-validation, and gap-test, are part of the publicly available GAP corpus. The preprocessing and sanitization steps are described next.

\begin{table}[t!]
\centering
\begin{tabular}{lllll}
     & M & F & Total\\
    \hline
    gap-development & 1000 & 1000 & 2000\\
    gap-validation & 227 & 227 & 454\\
    gap-test & 1000 & 1000 & 2000\\
    gpr-neither (section \ref{sec:neither}) & 129 & 124 & 253\\
    stage 2 test (kaggle)\textsuperscript{$\dagger$} & 6499 & 5860 & 12359
\end{tabular}
\caption{\label{table:corpus_stats}Corpus statistics. Masculine (M) and Feminine (F) instances were identified based on the gender of the pronoun mention labeled in the sample. \textsuperscript{$\dagger$}Only a subset of these may have been used for final evaluation.}
\end{table}

\subsection{Data Augmentation: \textit{Neither} instances}\label{sec:neither}
In an attempt to upsample and boost the classifier's confidence in the underrepresented $Neither$ category (Table \ref{table:label-sanitization}), about 250 instances were added manually. These were created by obtaining cluster predictions from the coreference model by \citeauthor{lee2018higher} \shortcite{lee2018higher} and choosing a pronoun and the two candidate entities A and B from disjoint clusters. However, in the interest of time, this strategy was not fully pursued. Instead, the evidence pooling module was used to resolve this problem, as will become clear from the discussion in section \ref{sec:discussions}.

\subsection{Mention Tags}\label{sec:mention_tags}
The raw text snippet is manipulated by enclosing the labeled span of mentions with their associated tags, i.e. $<$P$>$ for pronoun, $<$A$>$ for entity mention A, and $<$B$>$ for entity mention B. The primary reason for doing this is to provide the positional information of the labeled mentions implicitly within the text as opposed to explicitly through additional features. A secondary motivation was to test the language model's sensitivity to noise in input text structure, and its ability to adapt the pronoun representation to the positional tags. Figure \ref{fig:mention_tags} shows an example of this annotation scheme.

\begin{figure}[htb]
    \centering
    \small
    \begin{tabular}{p{180pt}}
        ... NHLer Gary Suter and Olympic-medalist \textbf{$<$A$>$} Bob Suter \textbf{$<$A$>$} are \textbf{$<$B$>$} Dehner \textbf{$<$B$>$}'s uncles. \textbf{$<$P$>$} His \textbf{$<$P$>$} cousin is Minnesota Wild's alternate captain Ryan ...\\
    \end{tabular}
    \caption{\label{fig:mention_tags}Sample text-snippet after annotating the mentions with their corresponding tags. Bob Suter and Dehner were tagged as entities A and B, and the mention 'His' following them was tagged as the pronoun. }
\end{figure}

\subsection{Label Sanitization}\label{sec:sanitization}
Only samples where labels can be corrected unambiguously based on snippet-context were corrected\footnote{Corrected labels can be found at \url{https://github.com/sattree/gap}. This set was generated independently to avoid any unintended bias. More sets of corrections can be found at \url{https://www.kaggle.com/c/gendered-pronoun-resolution/discussion/81331\#503094}}. The Wikipedia page-context and url-context were not used. A visualization tool \footnote{\url{https://github.com/sattree/gap/visualization}} was also developed as part of this work to aid in this activity. Table \ref{table:label-sanitization} lists the corpus statistics before and after the sanitization process.

\begin{table*}[t!]
\centering
\begin{tabular}{llllllll}
    \hline
     & \multicolumn{3}{l}{Before sanitization} & \multicolumn{3}{l}{After sanitization} & \\
    \cline{2-7}
     & A & B & NEITHER & A & B & NEITHER & Total\\
    \hline
    gap-development & 874	& 925 & 201 & 857(-37)(+20) & 919(-32)(+26)	& 224(-4)(+27) & 2000\\
    gap-validation & 187 & 205 & 62 & 184(-10)(+7) & 206(-7)(+8) & 64(-4)(+6) & 454\\
    gap-test & 918 & 855 & 227 & 894(-42)(+18) & 860(-27)(32) & 246(-8)(27) & 2000\\
\end{tabular}
\caption{\label{table:label-sanitization} GAP dataset label distribution before and after sanitization. (-x) indicates the number of samples that were moved out of a given class and (+x) indicates the number of samples that were added  post-sanitization.}
\end{table*}

\subsection{Coreference Signal}
Transformer networks have been found to have limited capability in modeling long-range dependency \cite{dai2018transformer, khandelwal2018sharp}. It has also been noticed in the past that the coreference problem benefits significantly from global knowledge \cite{lee2018higher}. Being cognizant of these two factors, it would be useful to inject predictions from off-the-shelf coreference models as an auxiliary source of evidence (with input text context being the primary evidence source). The models chosen for this purpose are Parallelism+URL \cite{webster2018mind}, AllenNLP\footnote{\url{https://allennlp.org/models}}, NeuralCoref\footnote{\url{https://github.com/huggingface/neuralcoref}}, and e2e-coref \cite{lee2018higher}.


\section{Model Architecture}

\subsection{ProBERT: baseline model}

ProBERT uses a fine-tuned BERT language model \cite{devlin2018bert, howard2018universal} with a classification head on top to serve as baseline. The snippet-text is augmented with mention-level tags (\cref{sec:mention_tags}) to capture the positional information of the pronoun, entity A, and entity B mentions, before feeding the text as input to the model. Position-wise token representation corresponding to the pronoun is extracted from the last layer of the language model. With GAP dataset and WordPiece tokenization \cite{devlin2018bert}, all pronouns were found to be single token entities. 

Let $E_p \in \mathbb{R}^H$ (where $H$ is the dimensionality of the language model output) denote the pooled pronoun vector. A linear transformation is applied to it, followed by softmax, to obtain a probability distribution over classes A, B, and NEITHER, $P=softmax(W^TE_p)$, where $W \in \mathbb{R}^{H \times 3}$ is the linear projection weight matrix. All the parameters are jointly trained to minimize cross entropy loss. This simple architecture is depicted in Figure \ref{probert}. Only $H \times 3$ new parameters are introduced in the architecture, allowing the model to use training data more efficiently. 

A natural question arises as to why this model functions so well (see \cref{sec:single_model_perf}) with just the pronoun representation. This is discussed in section \ref{sec:unreasonable_bert}.

\subsection{GREP: Gendered Resolution by Evidence Pooling}
The architecture for GREP pairs the simple ProBERT architecture with a novel Evidence Pooling module. The Evidence Pooling (EP) module leverages cluster predictions from pretrained (or heuristics-based) coreference models to gather evidence for the resolution task. The internals of the coreference models are opaque to the system, allowing for any evidence source such as a knowledge base to be included as well. This design choice limits us from propagating the gradients through the coreference models, thereby losing information and leaving them noisy. The difficulty of efficiently training deeper architectures paired with the noisy cluster predictions (the best coreference model has an $F1$ performance of only $64\%$ on gap-test) makes this a challenging design problem. The EP module uses self-attention mechanism described in \citet{vaswani2017attention} to compute the compatibility of cluster mentions with respect to the pronoun and the two candidates, entity A, and entity B. The simple and easily extensible architecture of this module is described next.

\begin{figure}[htbp]
  \centering
  \includegraphics[width=200pt]{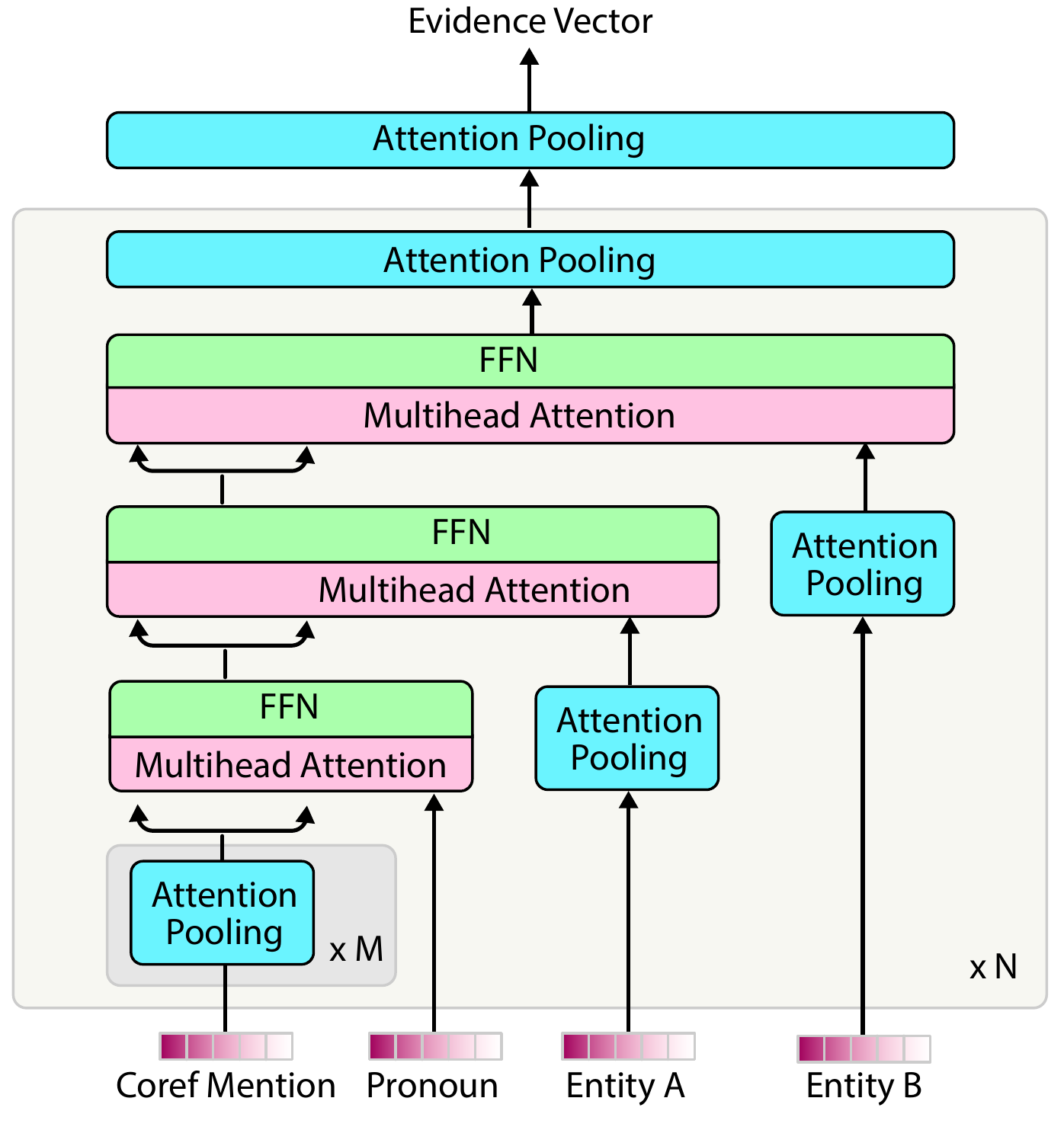}
  \caption{Evidence Pooling module architecture}
\end{figure}

Suppose we have access to N off-the-shelf coreference models and each predicts $T_n$ mentions that are coreferent with the given pronoun. Let P, A, and B, refer to the mentioned entities labeled in the text-snippet as the pronoun and entities A and B, respectively. Without loss of generality, let us consider the \textit{nth} coreference model and \textit{mth} mention in the cluster predicted by it. Let $E_m \in \mathbb{R}^{T_m \times H}$, $E_p \in \mathbb{R}^{T_p \times H}$, $E_a \in \mathbb{R}^{T_a \times H}$ and $E_b \in \mathbb{R}^{T_b \times H}$, denote the position-wise token embeddings obtained from the last layer of the language model for each of the mentions, where $T$ is the number of tokens in each mention. The first step is to aggregate the information at the mention-level. Self-attention is used to reduce the mention tokens, an operation that will be referred to as $AttnPool$ (attention pooling) hereafter. A single layer MLP is applied to compute position-wise compatibility score, which is then normalized and used to compute a weighted average over the mention tokens for a pooled representation of the mention as follows:

\begin{equation}
    M_m = tanh(E_mW_{m}+b_{m}) \in \mathbb{R}^{T_m \times H}
\end{equation}
\begin{equation}
    a_m = softmax(M_m) \in \mathbb{R}^{T_m}
\end{equation}
\begin{equation}
    AttnPool(E_m, W_m) = A_m = \sum_{i}^{T_m} a_mE_m \in \mathbb{R}^{H}
\end{equation}

Similarly, a pooled representation of all mentions in the cluster predicted by the $nth$ coreference model, and of P, A, and B entity mentions is obtained. Let $A_n \in \mathbb{R}^{T_n \times H}$ denote the joint representation of cluster mentions, and  $A_p$, $A_a$, and $A_b$, the pooled representations of entity mentions. Next, to compute the compatibility of the cluster with respect to the given entities, we systematically transform the cluster representation by passing it through a transformer layer \cite{vaswani2017attention}. A sequence of such transformations is applied successively by feeding $A_p$, $A_a$, and $A_b$ as query vectors at each stage. Each such transformer layer consists of a multi-head attention and feed-forward (FFN) projection layers. The reader is referred to \citet{vaswani2017attention} for further information on $MultiHead$ operation.

\begin{equation}
    FFN(x) = tanh(W_{x}x+b_{x}) \in \mathbb{R}^{T_p \times H}
\end{equation}
\begin{equation}
    C_p = FFN(MultiHead(A_p, A_m, A_m)) \in \mathbb{R}^{T_p \times H}
\end{equation}
\begin{equation}
    C_a = FFN(MultiHead(A_a, C_p, C_p)) \in \mathbb{R}^{T_a \times H}
\end{equation}
\begin{equation}
    C_b = FFN(MultiHead(A_b, C_a, C_a)) \in \mathbb{R}^{T_b \times H}
\end{equation}

The transformed cluster representation $C_b$ is then reduced at the cluster-level and finally at the coreference model level by attention pooling as:
\begin{equation}
    A_c = AttnPool(C_b, W_c) \in \mathbb{R}^{N \times H}
\end{equation}
\begin{equation}
    A_{co} = AttnPool(A_c, W_{co}) \in \mathbb{R}^H
\end{equation}

$A_{co}$ represents the evidence vector that encodes information obtained from all the coreference models. Finally, the evidence vector is concatenated with the pronoun representation, and is once again fed through a linear layer and softmax to obtain class probabilities.

\begin{equation}
    C = [E_p;A_{co}] \in \mathbb{R}^{2H}
\end{equation}
\begin{equation}
    P = softmax(W^TC+b) \in \mathbb{R}^{3}
\end{equation}

\begin{figure}[htbp]
  \centering
  \includegraphics[width=200pt]{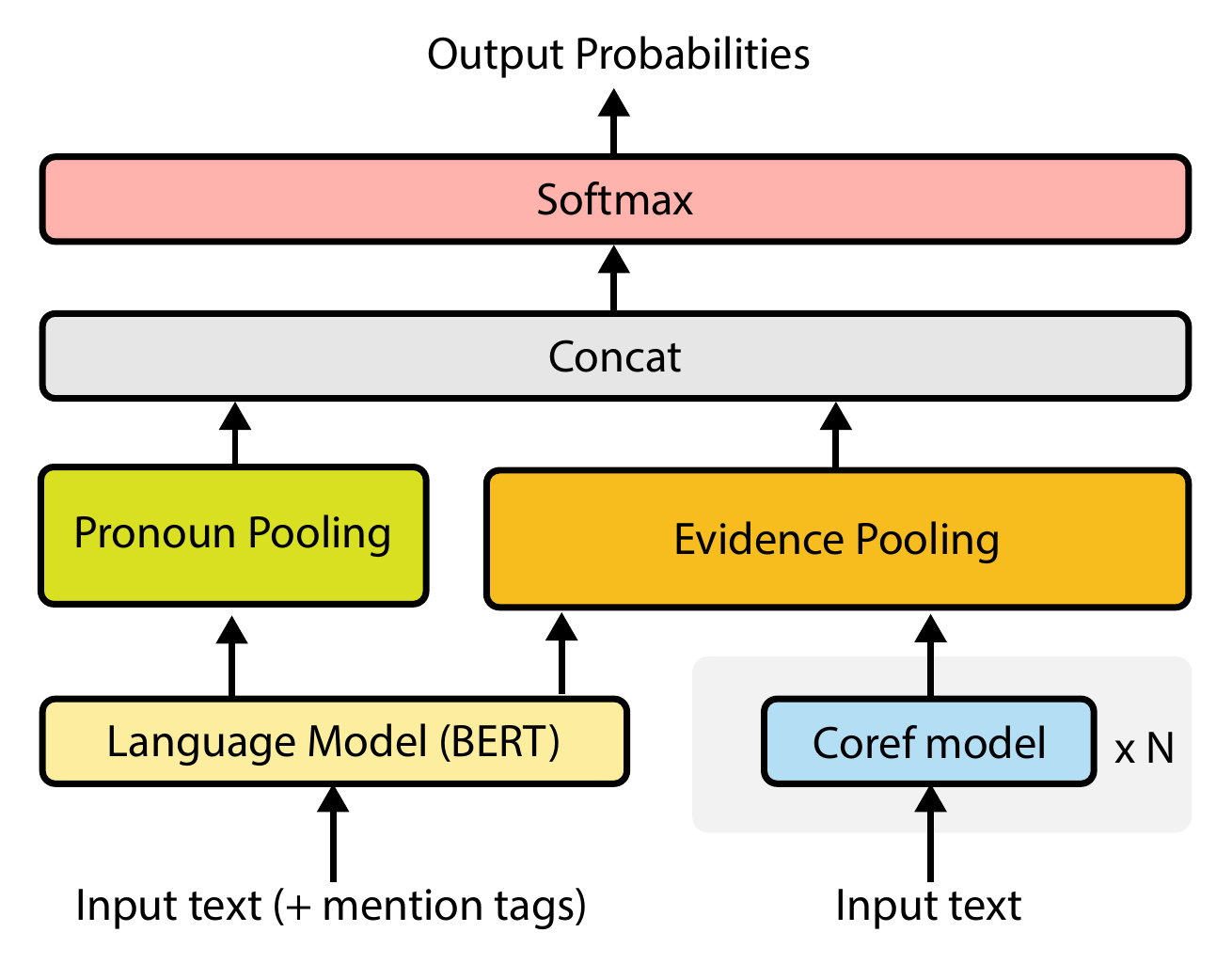}
  \caption{\label{fig:grep}GREP model architecture}
\end{figure}

The end-to-end GREP model architecture is illustrated in Figure \ref{fig:grep}.

\begin{table*}[t!]
\centering
\begin{tabular}{llllll}
     & \multicolumn{4}{c}{F1} & \multirow{2}{*}{logloss}\\
     \cline{2-5}
     & M & F & B & O & \\
    \hline
    Lee et al. (2017)\textsuperscript{$\dagger$} & 67.7 & 60.0 & 0.89 & 64.0 & -\\
    Parallelism\textsuperscript{$\dagger$} & 69.4 & 64.4 & 0.93 & 66.9 & -\\
    Parallelism+URL\textsuperscript{$\dagger$} & 72.3 & 68.8 & 0.95 & 70.6 & -\\
    RefReader, LM \& coref\textsuperscript{$\ddagger$} & 72.8 & 71.4 & \textbf{0.98} & 72.1 & -\\
    \hline
    ProBERT (bert-base-uncased) & 88.9 & 86.7 & \textbf{0.98} & 87.8 & .382\\
    GREP (bert-base-uncased) & 90.4 & 87.6 & 0.97 & 89.0 & .350\\
    \makecell{ProBERT (bert-large-uncased)} & 90.8	& 88.6 & \textbf{0.98} & 89.7 & .376\\
    \makecell{\textbf{GREP} \textbf{(bert-large-uncased)}} & \textbf{94.0} & \textbf{91.1}	& 0.97 & \textbf{92.5} & \textbf{.317}\\
    \hline
    Human Performance (estimated) & 97.2 & 96.1 & 0.99 & 96.6 & -\\
\end{tabular}
\caption{\label{table:single_model_perf}Single model performance on gap-test set by gender. M: masculine, F: feminine, B: (bias) ratio of feminine to masculine performance, O: overall. Log loss is not available for systems that only produce labels. \textsuperscript{$\dagger$}As reported by \citeauthor{webster2018mind} \shortcite{webster2018mind}. \textsuperscript{$\ddagger$}As reported by \citeauthor{liu2019referential} \shortcite{liu2019referential}, their model does not use \textit{gold-two-mention} labeled span information for prediction.}
\end{table*}

\section{Training} \label{sec:training}
All models were trained on 4 NVIDIA V100 GPUs (16GB memory each). The \textit{pytorch-pretrained-bert}\footnote{\url{https://github.com/huggingface/pytorch-pretrained-BERT/}. BertTokenizer from this package was used for tokenization of the text. BertAdam was used as the optimizer. This package contains resources for all variants of BERT, i.e. bert-base-uncased, bert-base-cased, bert-large-uncased and bert-large-cased.} library was used as the language model module and saved model checkpoints were used for initialization. Adam \cite{kingma2014adam} optimizer was used with $\beta_1=0.9$, $\beta_2=0.999$, $\epsilon=1e^{-6}$, and a fixed learning rate of $4e^{-6}$. For regularization, a fixed dropout \cite{srivastava2014dropout} rate of 0.1 was used in all layers and a weight decay of 0.01 was applied to all parameters. Batch sizes of 16 and 8 samples were used for model variants with bert-base and bert-large respectively. Models with bert-base took about 6 mins to train while those with bert-large took up to 20 mins. 

For single model performance evaluation, the models were trained on gap-train, early-stopping was based off of gap-validation, and gap-test was used for test evaluation. Kaggle competition results were obtained by training models on all datasets, i.e. gap-train, gap-validation, gap-test, and gpr-neither (a total of 4707 samples), in a 5-Fold Cross-Validation \cite{friedman2001elements} fashion. Each model gets effectively trained on 3768 samples, while 942 samples were held-out for validation. Training would terminate upon identifying an optimal early stopping point based on performance on the validation set with an evaluation frequency of 80 gradient steps. Model's access is limited to snippet-context, and the Wikipedia page-context is not used. However, page-url context may be used via coreference signal (Parallelism+URL).

\section{Results}\label{sec:results}

\begin{table*}[t!]
\centering
\begin{tabular}{lllllll}
     \multirow{2}{*}{Model} & \multirow{2}{*}{Dataset} & \multicolumn{4}{c}{F1} & \multirow{2}{*}{logloss}\\
     \cline{3-6}
      & & M & F & B & O & \\
    \hline
    \multirow{2}{*}{LM=bert-large-uncased, seed=42} & OOF all & 94.3 & 93.21 & 0.99 & 93.8 & .261\\
    & OOF gap-test & 94.2 & 93.7 & 0.99 & 93.9 & .254\\
    \hline
    \multirow{2}{*}{LM=bert-large-cased, seed=42} & OOF all  & 94.3	& 93.9 & 0.99 & 94.1 & .249\\
    & OOF gap-test & 94.3 & 93.5 & 0.99 & 93.9 & .242\\
    \hline
    \multirow{2}{*}{Ensemble: \makecell{(LM=bert-large-uncased \\+ seeds=42,59,75,46,91)}} & OOF all  & 94.8 & 94.2 & 0.99 & 94.5 & .195\\
    & OOF gap-test & 94.5 & 94.33 & 1.00 & 94.4 & .193\\
    \hline
    \multirow{2}{*}{Ensemble: \makecell{(LM=bert-large-cased \\+ seeds=42,59,75,46,91)}} & OOF all & 95.1 & 94.4 & 0.99 & 94.7 & .187\\
    & OOF gap-test & 94.9 & 94.1 & 0.99 & 94.4 & .183\\
    \hline
    \multirow{3}{*}{Ensemble: \makecell{(LM=bert-large-uncased, \\ \qquad\quad bert-large-cased \\+ seeds=42,59,75,46,91)}} & OOF all & 95.3 & 94.7 & 0.99 & 95.0 & .176\\
    & OOF gap-test & 95.1 & 94.7 & 1.00 & 94.9 & .175\\
    \cline{2-7}
    & \textbf{Stage 2 test} & - & - & - & - & \textbf{.137\textsuperscript{$\dagger$}}\\
\end{tabular}
\caption{\label{table:kaggle_resuts}GREP model performance results in  the Kaggle competition. Out-of-fold (OOF) error is reported on all data, i.e. gap-development, gap-validation, gap-test, and gpr-neither, as well as on gap-test explicitly for comparison against single model performance results. Since early stopping is based on OOF samples, OOF errors reported here cannot be considered as an estimate of test error. Nevertheless, stage 2 test performance benchmarks the model. \textsuperscript{$\dagger$}Due to a bug, the model did not fully leverage coref evidence, further gains are expected with the fixed version.}
\end{table*}

The performance of ProBERT and GREP models is benchmarked against results previously established by  \citeauthor{webster2018mind} and \citeauthor{liu2019referential} \shortcite{liu2019referential}. It is worth noting that \citeauthor{liu2019referential} do not use \textit{gold-two-mention} labeled spans for prediction and hence their results may not be directly comparable.  This section first introduces an estimate of human performance on this task. Then, results for single model performance are presented, followed by ensembled model results that won the Kaggle competition. $F1$ performance scores were obtained by using the GAP scorer script\footnote{\url{https://github.com/google-research-datasets/gap-coreference/blob/master/gap\_scorer.py}} provided by \citeauthor{webster2018mind}. Wherever applicable, log loss (the official Kaggle metric) performance is reported as well.

\subsection{Human Performance} Errors found in crowd-sourced labels are considered a measure of human performance on this task, and serve as a benchmark. The corrections are only a best-effort attempt to fix some obvious mistakes found in the dataset labels, and were made with certain considerations (section \ref{sec:sanitization}). This performance measure is subject to variation based on an evaluator's opinion on ambiguous samples.

\subsection{Single Model Performance} \label{sec:single_model_perf}
Single model performance on GAP test set is shown in Table \ref{table:single_model_perf}. The GREP model (with bert-large-uncased as the language model) achieves a powerful state-of-the-art performance on this task. The model significantly benefits from evidence pooling, gaining 6 points in terms of log loss and 2.8 points in $F1$ accuracy. Further analysis of the source of these gains is discussed in section \ref{sec:discussions}. 

While it may seem that the significantly improved performance of GREP has been achieved at a small cost in terms of gender bias, an attentive reader would realize that the model enjoys improved performance for both genders. Performance gains in masculine instances are much higher compared to feminine instances, and the slight degradation in bias ratio is a manifestation of this. The superior performance of GREP provides evidence that for a given sample context, the model architecture is able to successfully discriminate between the coreference signals, and identify their usefulness.


\subsection{Kaggle Competition\footnote{\url{https://www.kaggle.com/c/gendered-pronoun-resolution/}}}

To encourage fairness in modeling, the competition was organized in two stages. This strategy eliminates any attempts at leaderboard probing and other such malpractices. Furthermore, models were frozen at the end of stage 1 and were only allowed to operate in inference mode to generate predictions for stage 2 test submission. Additionally, no feedback was provided on stage 2 submissions (in terms of performance score) until the end of the competition.

 GREP model is trained as described in section \ref{sec:training} and out-of-fold (oof) error on the held-out samples is reported. The experiments are repeated with 5 different random seeds (42, 59, 75, 46, 91) for initialization. Finally, two sets of models are trained with bert-large-uncased and bert-large-cased as the language models. The overall scheme leads to 50 models being trained in total, and 50 sets of predictions being generated on stage 2 test data. To generate predictions for submission, ensembling is done by simply taking the unbiased weighted mean over the 50 individual prediction sets. 
 
 Table \ref{table:kaggle_resuts} presents a granular view of the winning model performance. This performance comes very close to human performance and has almost no gender bias. As the table shows, the ensemble models achieve much larger gains in log loss as compared to $F1$ accuracy. This is expected since the committee of models makes more confident decisions on ``easier" examples. Two insights can be drawn by comparing these results with the single model performance presented in section \ref{sec:single_model_perf}: (1) model accuracy benefits from more training data, although the gains are marginal at best (92.5 vs 93.9) given that the model was trained on approximately twice the amount of data; (2) ensembling has a similar effect as evidence pooling, i.e., models become more confident in their predictions. 

\begin{figure*}[htbp]
     \centering
     \subfloat[A \label{fig:a}]{
         \centering
         \includegraphics[width=0.28\textwidth]{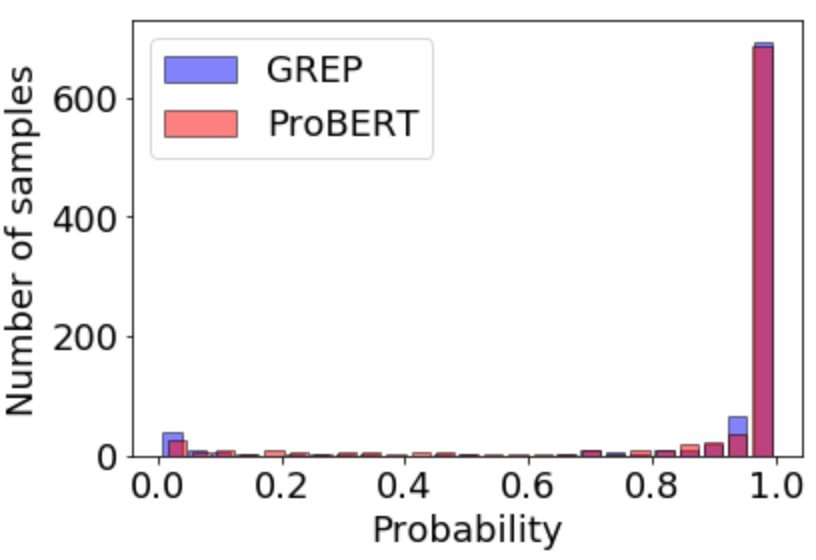}
         }
     \hfill
     \subfloat[B\label{fig:b}]{
         \centering
         \includegraphics[width=0.28\textwidth]{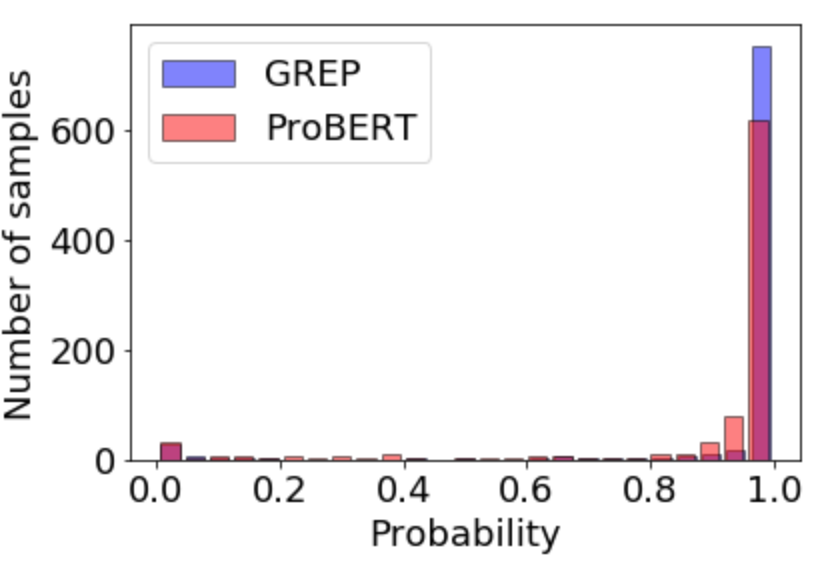}
         }
     \hfill
     \subfloat[NEITHER\label{fig:c}]{
         \centering
         \includegraphics[width=0.28\textwidth]{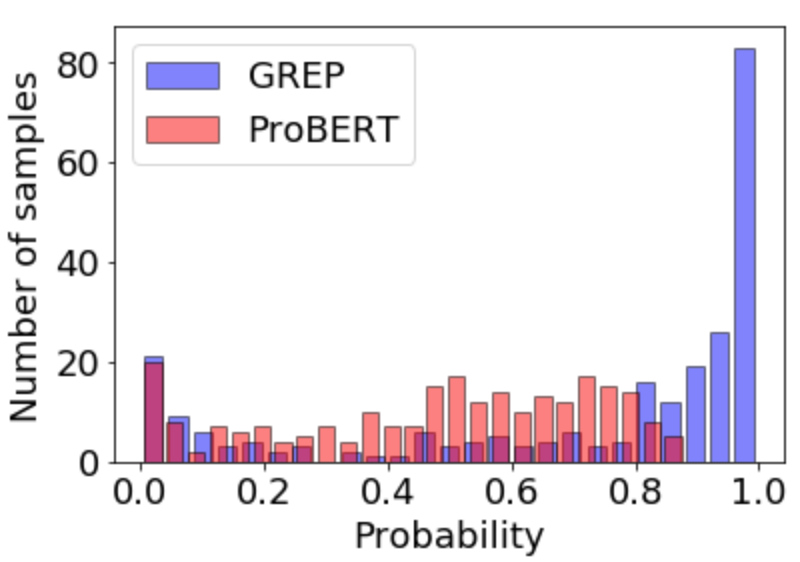}
         }
    \caption{Comparison of probabilities assigned by ProBERT and GREP. Figures show distribution of predicted class probabilities assigned by the models to samples from that class.}
        \label{fig:probabilities}
\end{figure*}

\section{Discussion}\label{sec:discussions}
Results shown in section \ref{sec:results} establish the superior performance of GREP compared to ProBERT. This can be attributed to two sources: (1) GREP corrects some errors made by ProBERT, reflected in F1; and (2) where predictions are correct, GREP is more confident in its predictions, reflected in log loss. To investigate this, error analysis is performed on gap-test.

Figure \ref{fig:probabilities} shows a class-wise comparison of probabilities generated by the two models. It can be seen that GREP is more confident in its predictions (all distributions appear translated closer to 1.0), and the improvement is overwhelmingly evident for the NEITHER class. To understand the difference between the two models, confusion matrix statistics are presented in table \ref{table:confusion}. The diagonal terms show the number of instances that the two models agree on, and the off-diagonal terms show where they disagree. The numbers reveal that the evidence pooling module not only boosts the model confidence but also helps in correctly resolving \textit{Neither} instances (44 vs 11), indicating that the model is successfully able to build evidence for or against the given candidates.

\begin{table}[t!]
\centering
\begin{tabular}{llll}
    & & \multicolumn{2}{c}{GREP}\\
    \cline{3-4}
     & ProBERT  & Incorrect & Correct\\
    \hline
    \multirow{2}{*}{A} & Incorrect & 44 & 38\\
                             & Correct & 28 & 784\\
    \hline
    \multirow{2}{*}{B} & Incorrect & 37 & 39\\
                             & Correct & 9 & 775\\
    \hline
    \multirow{2}{*}{NEITHER} & Incorrect & 45 & 44\\
                             & Correct & 11 & 146\\
    \hline
    \multirow{2}{*}{Overall} & Incorrect & 126 & 121\\
                             & Correct & 48 & 1705\\
\end{tabular}
\caption{\label{table:confusion}Class-wise comparison of model accuracy for ProBERT and GREP. Off-diagonal terms show cases where GREP fixes errors made by ProBERT and vice-versa.}
\end{table}

Appendix \ref{sec:appendix} details the behavior of GREP through some examples. The first example is particularly interesting - while it is trivial for a human to resolve this, a machine would require knowledge of the world to understand ``death" and its implications.

\subsection{Unreasonable Effectiveness of ProBERT}
\label{sec:unreasonable_bert}
It would seem unreasonable that ProBERT is able to perform so well with the noisy input text (due to mention tags) and is able to make the classification decision by looking at the pronoun alone. The following two theories may explain this behavior: (1) attention heads in the (BERT) transformer architecture are able to specialize the pronoun representation in the presence of the supervision signal; (2) the special nature of dropout (present in every layer) makes the model immune to a small amount of noise, and at the same time prevents the model from ignoring the tags. The analysis of attention heads to investigate these claims should form the scope of future work.

\section{Conclusion}
A powerful set of results have been established for the shared task. Work presented in this paper makes it feasible to efficiently employ neural attention for pooling information from auxiliary sources of global knowledge. The evidence pooling mechanism introduced here is able to leverage upon the strengths of off-the-shelf coreference solvers without being hindered by their weaknesses (gender bias). A natural extension of the GREP model would be to solve the gendered pronoun resolution problem beyond the scope of the \textit{gold-two-mention} task, i.e., without accessing the labeled gold spans.

\section*{Acknowledgments}
I would like to thank Google AI Language and Kaggle for hosting and organizing this competition, and for providing a platform for independent research. 

\bibliography{acl2019}
\bibliographystyle{acl_natbib}

\appendix

\section{Examples}
\label{sec:appendix}
\begin{figure*}[htbp]
    \subfloat[Coreference model level attention weights. Indicates weightage given to evidence from each source.]{
        \includegraphics[width=\textwidth]{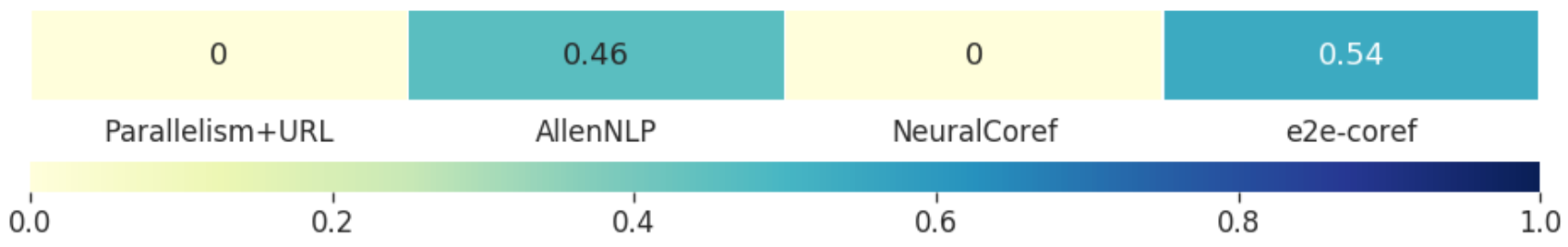}
    }
    \vfill
    \vspace{1cm}
    \subfloat[Cluster mention level attention weights. Indicates weightage given to each mention within an evidence cluster.]{
        \includegraphics[width=\textwidth]{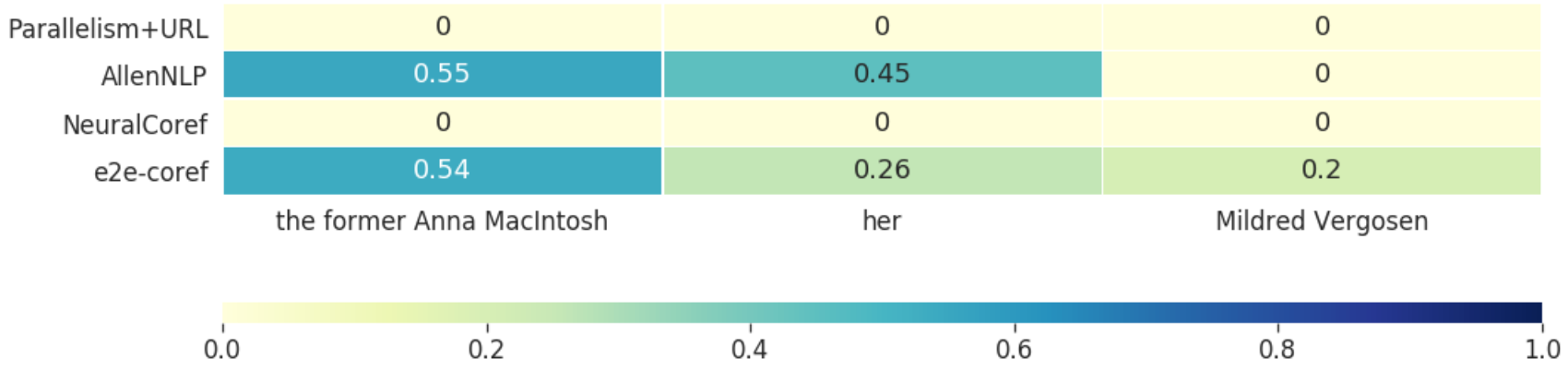}
    }
    \caption{Example 1 - Visualization of normalized attention scores assigned by the hierarchical attention pooling layers in the evidence pooling module}
    \label{fig:example1attn}
\end{figure*}

Tables \ref{table:example1vis}, \ref{table:example1sample}, \ref{table:example1probs}, and Figure \ref{fig:example1attn} show an example of how incorporating evidence from the coreference models helps GREP to correct a prediction error made by ProBERT. While the example is trivial for a human to resolve, a machine would require knowledge of the world to understand ``death" and its implications. ProBERT is unsure about the resolution and ends up assigning comparable probabilities to both entities A and B. GREP, on the other hand, is able to shift nearly all the probability mass from B to the correct resolution A, in light of strong evidence presented by the coreference solvers. Figure \ref{fig:example1attn} illustrates an interesting phenomenon; while e2e-coref groups the pronoun and both entities A and B in the same cluster, the model architecture is able to harvest information from AllenNLP predictions, propagating the belief that entity A must be the better candidate. The above observations indicate that by pooling evidence from various sources, the model is able to reason over a larger space and build a rudimentary form of world knowledge.

\begin{figure*}[!htpb]
    \subfloat[Coreference model level attention weights. Indicates weightage given to evidence from each source.]{
        \includegraphics[width=\textwidth]{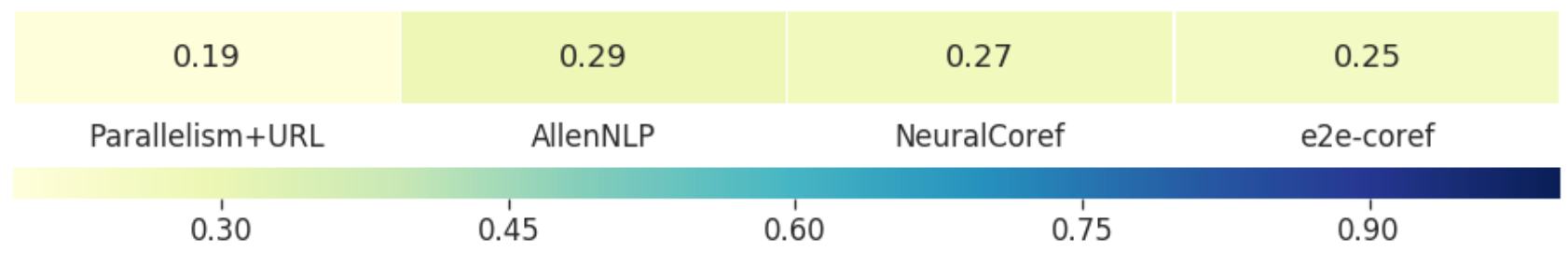}
    }
    \vfill
    \vspace{1cm}
    \subfloat[Cluster mention level attention weights. Indicates weightage given to each mention within an evidence cluster.]{
        \includegraphics[width=\textwidth]{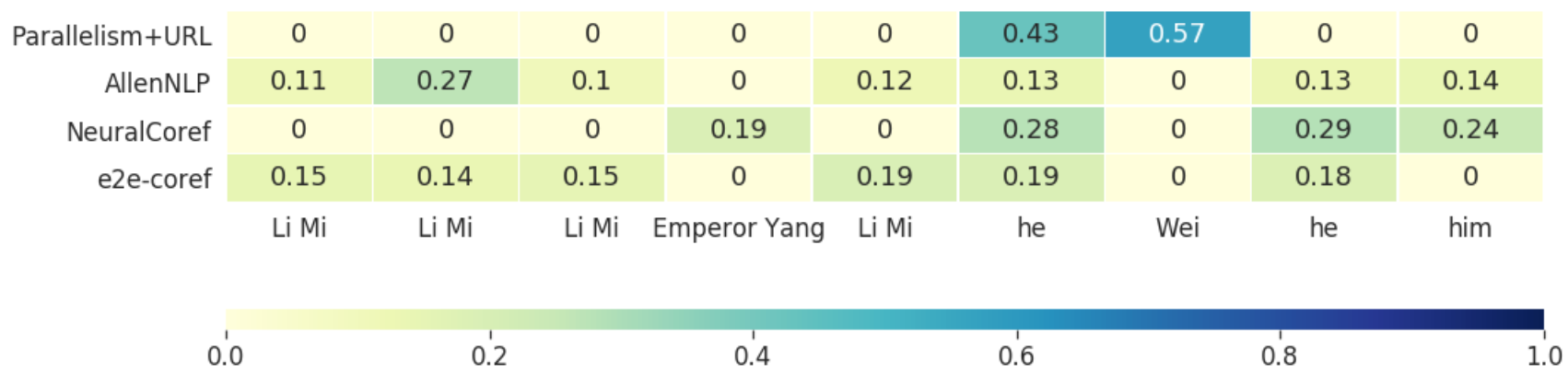}
    }
    \caption{Example 2 - Visualization of normalized attention scores assigned by the hierarchical attention pooling layers in the evidence pooling module}
    \label{fig:example2attn}
\end{figure*}

Tables \ref{table:example2vis}, \ref{table:example2sample}, \ref{table:example2probs}, and Figure \ref{fig:example2attn} show a second example. This example is not easy even for a human to resolve without reading and understanding the full context. A model may find this situation to be adverse given the presence of too many named entities as distractor elements; and the url-context can be misleading since the pronoun referent is not the subject of the article. Nevertheless, the model is able to successfully build evidence against the given candidates, and resolve with a very high confidence of 92.5\%.

Finally, a third example is shown in Tables \ref{table:example3vis} and \ref{table:example3probs}. This example shows that the model doesn't simply make a majority decision, rather considers interactions between the global structure exposed by the various evidence sources.

\begin{table*}[t!]
\centering
\begin{tabular}{l}
    \hline
    \\[-1em]
    \textbf{Ground truth}\\
    \hline
    \\[-1em]
    \includegraphics[width=\textwidth]{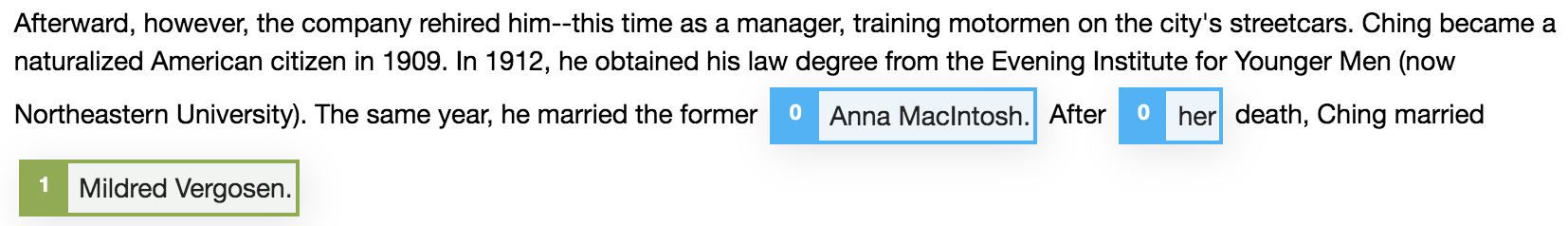}
    \\[-1em]
    \\
    \hline
    \\[-1em]
    \textbf{Off-the-shelf coreference model predictions}\\
    \\[-1em]
    \hline
    \\[-1em]
    \subfloat[Parallelism+URL]{
        \includegraphics[width=\textwidth]{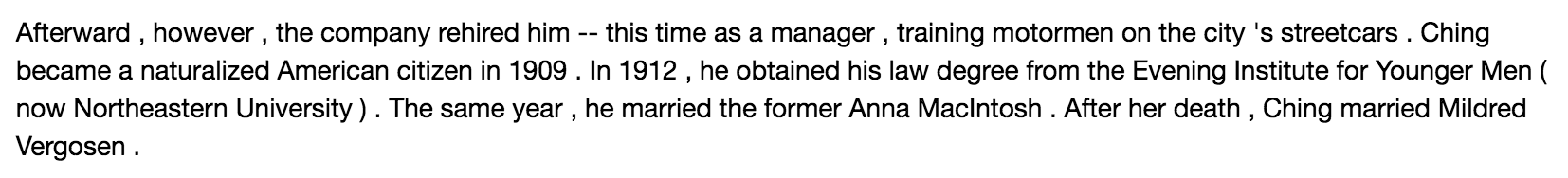}
    }
    \\
    \\
    \subfloat[AllenNLP]{
        \includegraphics[width=\textwidth]{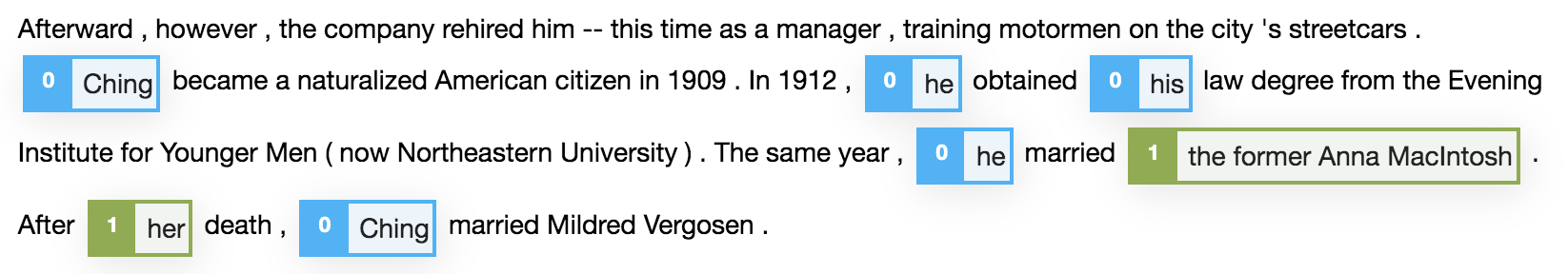}
    }
    \\
    \\
    \subfloat[NeuralCoref]{
        \includegraphics[width=\textwidth]{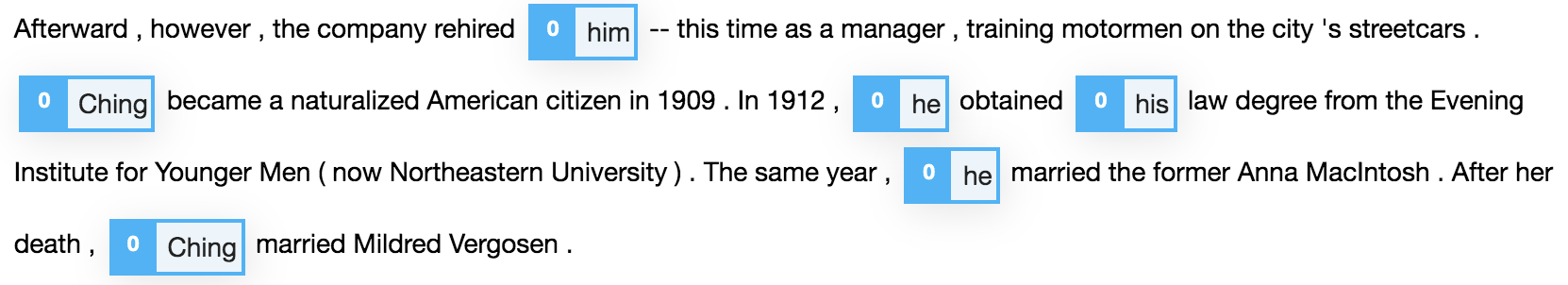}
    }
    \\
    \\
    \subfloat[e2e-coref \cite{lee2018higher}]{
        \includegraphics[width=\textwidth]{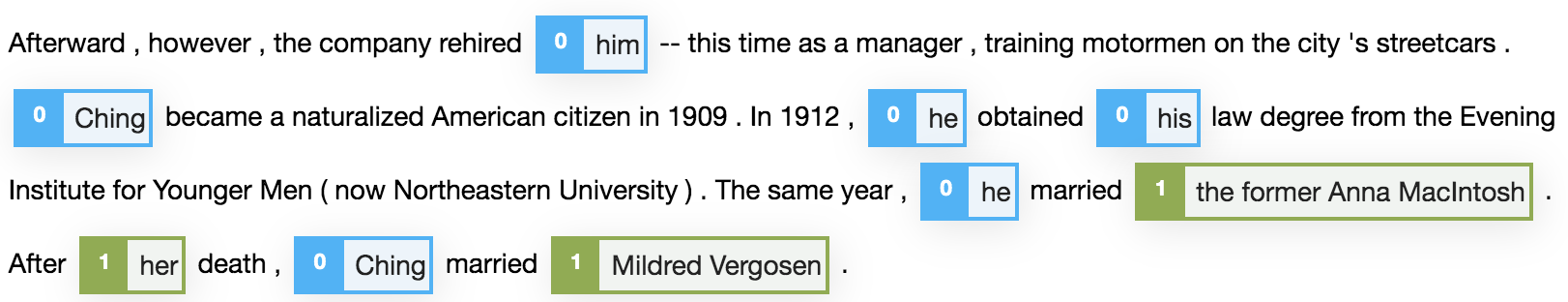}
    }
    \\
\end{tabular}
\caption{\label{table:example1vis} Example 1 - Illustration of ground truth and coreference model predictions. Mentions belonging to a coreference cluster are color coded and indexed. Visualizations were produced using the code module at \url{https://github.com/sattree/gap/visualization}.}
\end{table*}

\FloatBarrier

\begin{table*}[t!]
\centering
    \begin{tabular*}{2\columnwidth}{@{\extracolsep{\fill}}llllllllll}
        \hline
        \makecell{id \\ \quad}	& \makecell{Pronoun \\ \quad} &	\makecell{Pronoun \\offset}	& \makecell{A \\ \quad} & \makecell{A \\offset} &  \makecell{A \\ coref}	& \makecell{B \\ \quad} & \makecell{B \\offset}	& \makecell{B \\ coref} & \makecell{Url \\ \quad}\\ 
        \hline
        \makecell{test-\\282} & her & 410 & \makecell{Anna \\MacIntosh} &	338 & True &	\makecell{Mildred \\Vergosen} &	475 & False & \makecell*[{{p{2cm}}}]{\url{http://en.wikipedia.org/wiki/Cyrus_S._Ching}}  \\ \hline
    \end{tabular*} \\ 
    \caption{\label{table:example1sample} Example 1 - Sample details from GAP test set.}
\end{table*}

\begin{table*}[t!]
\centering
\begin{tabular*}{2\columnwidth}{@{\extracolsep{\fill}}lccc}
    \hline
         & P(A) & P(B) & P(NEITHER)\\ 
        \hline
        ProBERT & 0.405 & 0.452 & 0.142  \\ 
        GREP & 0.718 & 0.038 & 0.244\\
    \hline
\end{tabular*}
\caption{\label{table:example1probs} Example 1 - A comparison of probabilities assigned by ProBERT and GREP}
\end{table*}

\begin{table*}[t!]
\centering
\begin{tabular}{l}
    \hline
    \\[-1em]
    \textbf{Ground truth}\\
    \hline
    \\[-1em]
    \includegraphics[width=\textwidth]{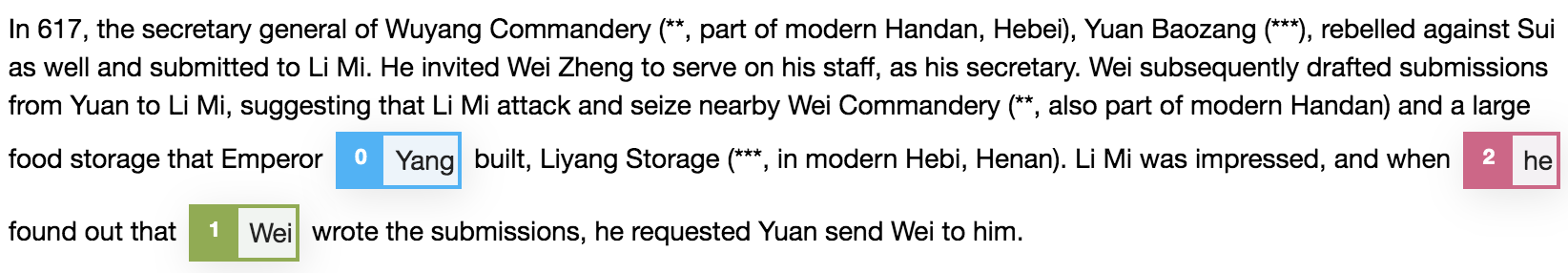}
    \\[-1em]
    \\
    \hline
    \\[-1em]
    \textbf{Off-the-shelf coreference model predictions}\\
    \\[-1em]
    \hline
    \\[-1em]
    \subfloat[Parallelism+URL]{
        \includegraphics[width=\textwidth]{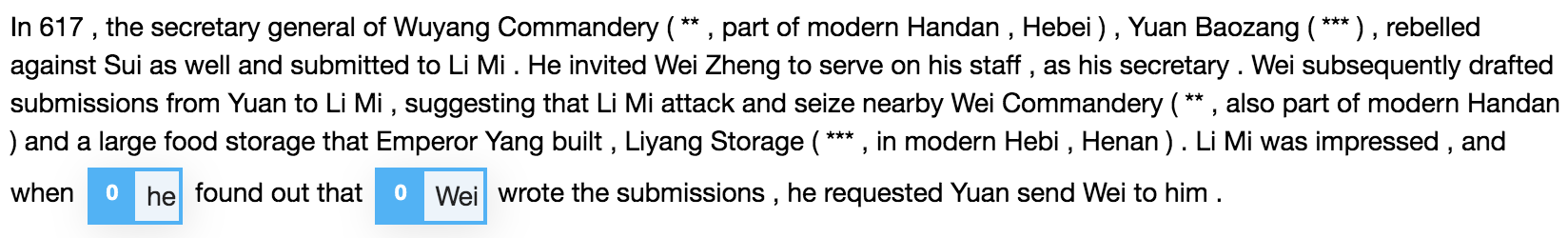}
    }
    \\
    \\
    \subfloat[AllenNLP]{
        \includegraphics[width=\textwidth]{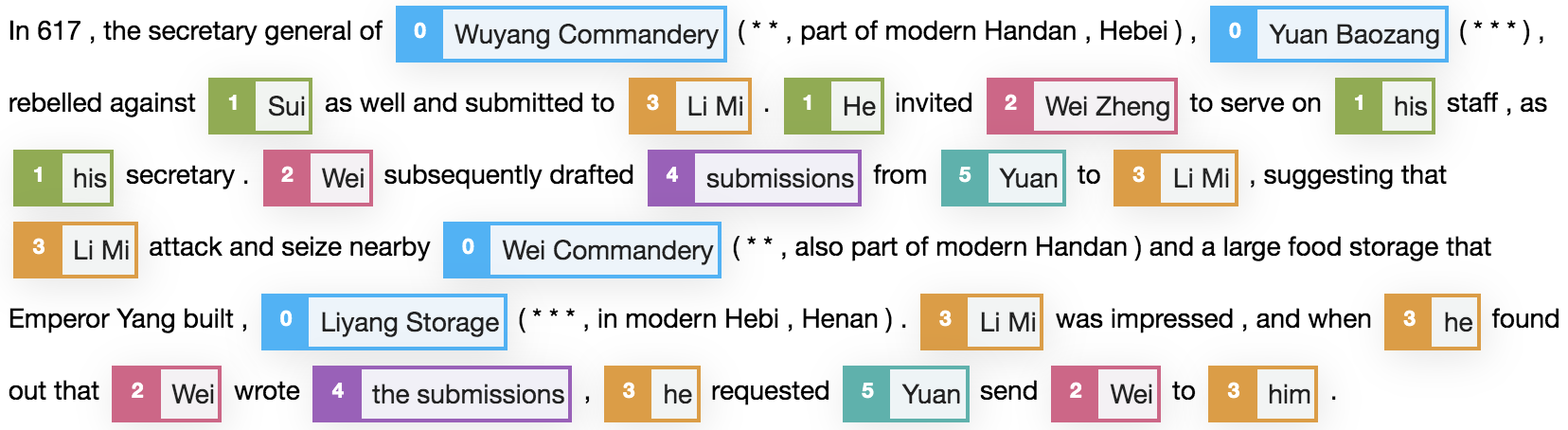}
    }
    \\
    \\
    \subfloat[NeuralCoref]{
        \includegraphics[width=\textwidth]{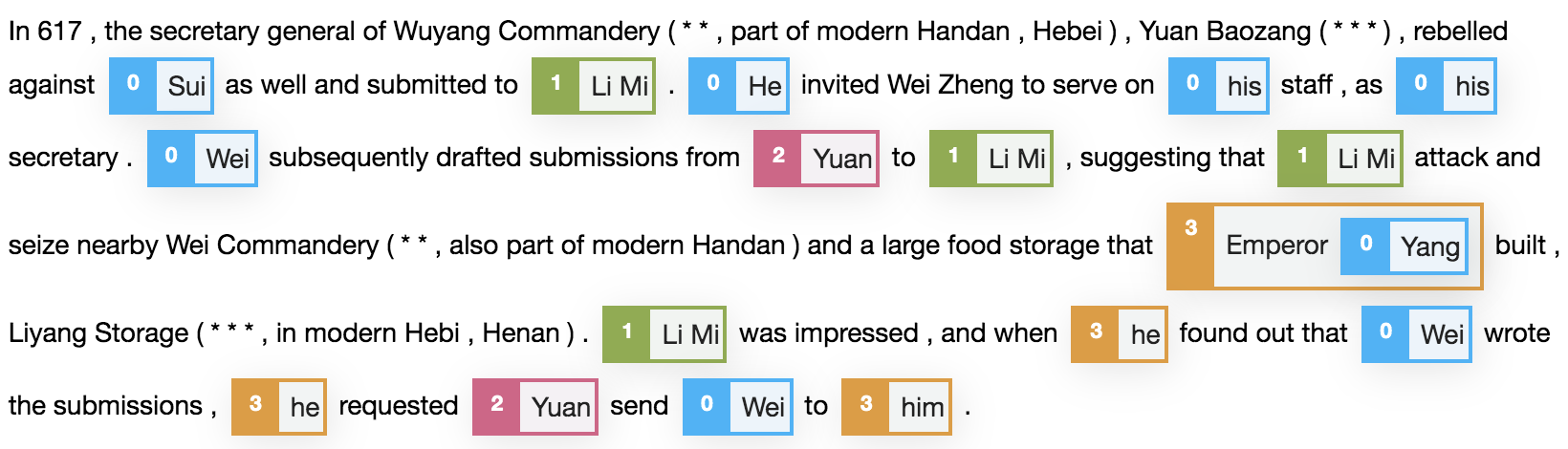}
    }
    \\
    \\
    \subfloat[e2e-coref \cite{lee2018higher}]{
        \includegraphics[width=\textwidth]{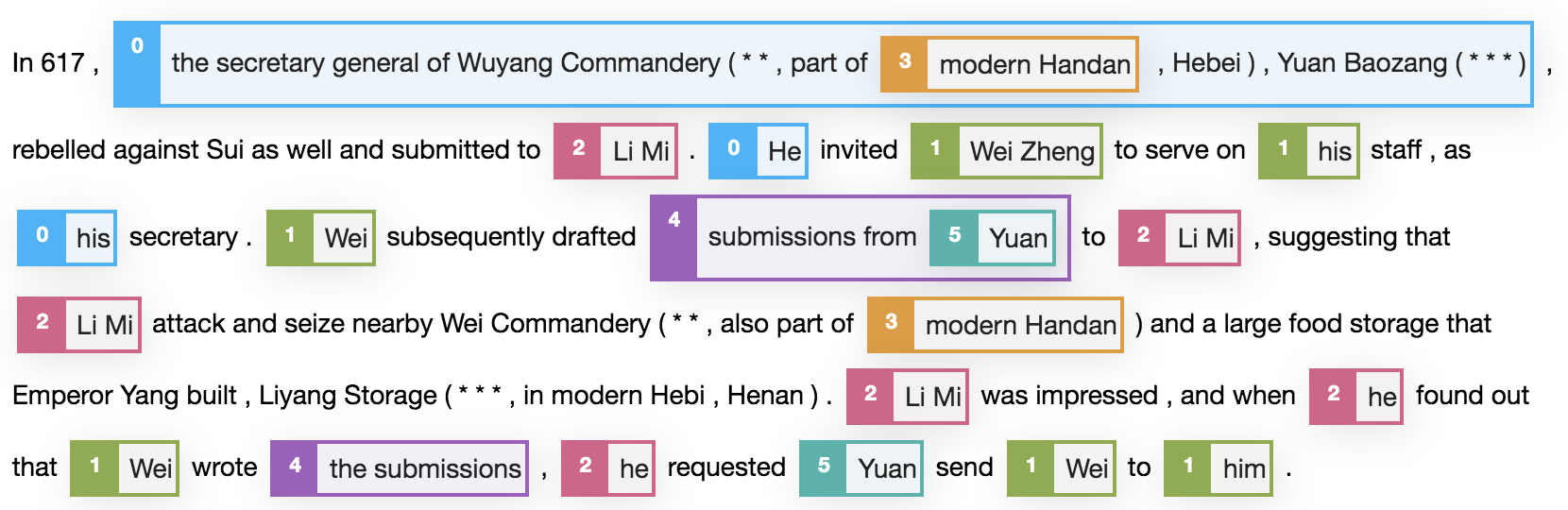}
    }
    \\
\end{tabular}
\caption{\label{table:example2vis} Example 2 - Illustration of ground truth and coreference model predictions. Mentions belonging to a coreference cluster are color coded and indexed.}
\end{table*}

\FloatBarrier

\begin{table*}[t!]
\centering
    \begin{tabular*}{2\columnwidth}{@{\extracolsep{\fill}}llllllllll}
        \hline
        \makecell{id \\ \quad} & \makecell{Pronoun \\ \quad} & \makecell{Pronoun \\ offset} & \makecell{A \\ \quad} & \makecell{A \\ offset} & \makecell{A \\ coref} & \makecell{B \\ \quad} & \makecell{B \\ offset} & \makecell{B \\ coref} & \makecell{Url \\ \quad} \\
        \hline
        \makecell{test-\\406} & he & 803 & Yang & 636 & False & Wei & 916 & False & \makecell*[{{p{2cm}}}]{\url{http://en.wikipedia.org/wiki/Wei_Zheng}}\\
        \hline
    \end{tabular*} \\ 
    \caption{\label{table:example2sample} Example 2 - Sample details from GAP test set.}
\end{table*}

\begin{table*}[t!]
\centering
\begin{tabular*}{2\columnwidth}{@{\extracolsep{\fill}}lccc}
    \hline
         & P(A) & P(B) & P(NEITHER)\\ 
        \hline
        ProBERT & 0.790 & 0.038 & 0.172  \\ 
        GREP & 0.055 & 0.020 & 0.925\\
    \hline
\end{tabular*}
\caption{\label{table:example2probs} Example 2 - A comparison of probabilities assigned by ProBERT and GREP}
\end{table*}

\begin{table*}[t!]
\centering
\begin{tabular}{l}
    \hline
    \\[-1em]
    \textbf{Ground truth}\\
    \hline
    \\[-1em]
    \includegraphics[width=\textwidth]{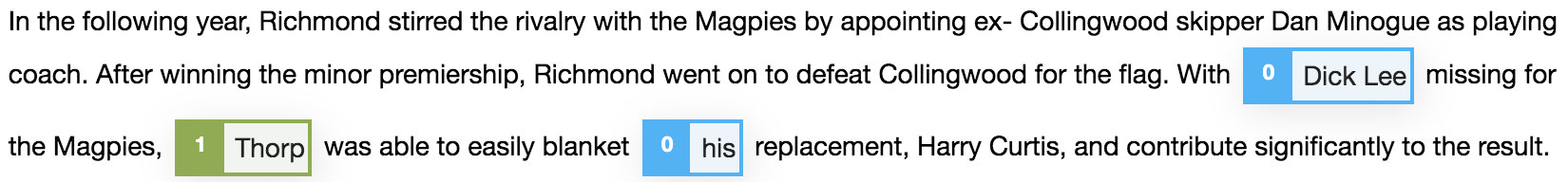}
    \\[-1em]
    \\
    \hline
    \\[-1em]
    \textbf{Off-the-shelf coreference model predictions}\\
    \\[-1em]
    \hline
    \\[-1em]
    \subfloat[Parallelism+URL]{
        \includegraphics[width=\textwidth]{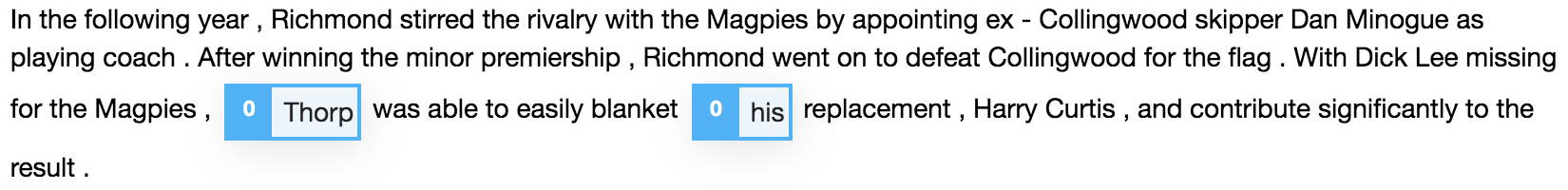}
    }
    \\
    \\
    \subfloat[AllenNLP]{
        \includegraphics[width=\textwidth]{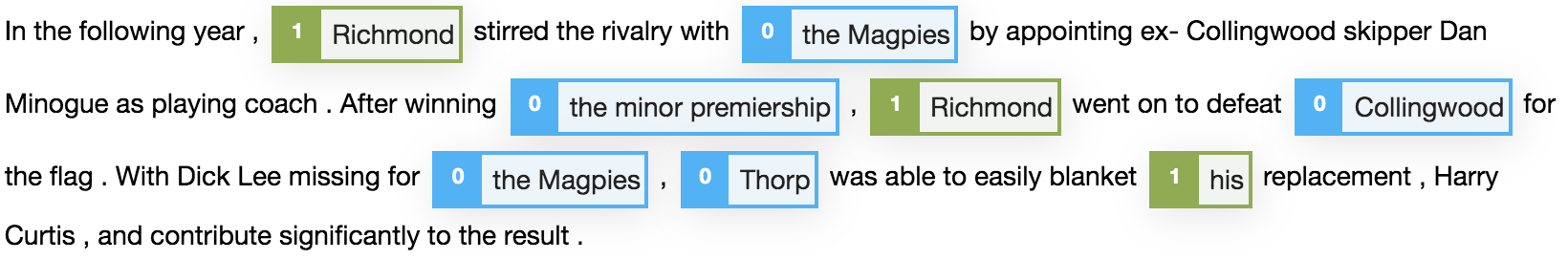}
    }
    \\
    \\
    \subfloat[NeuralCoref]{
        \includegraphics[width=\textwidth]{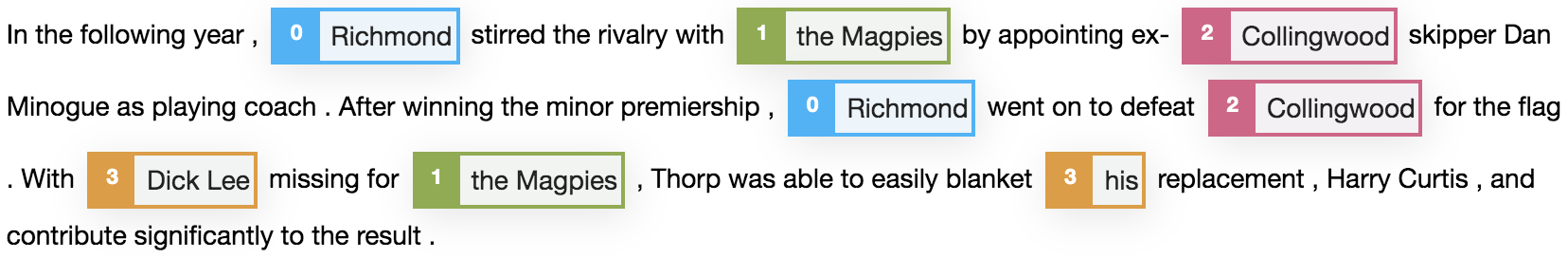}
    }
    \\
    \\
    \subfloat[e2e-coref \cite{lee2018higher}]{
        \includegraphics[width=\textwidth]{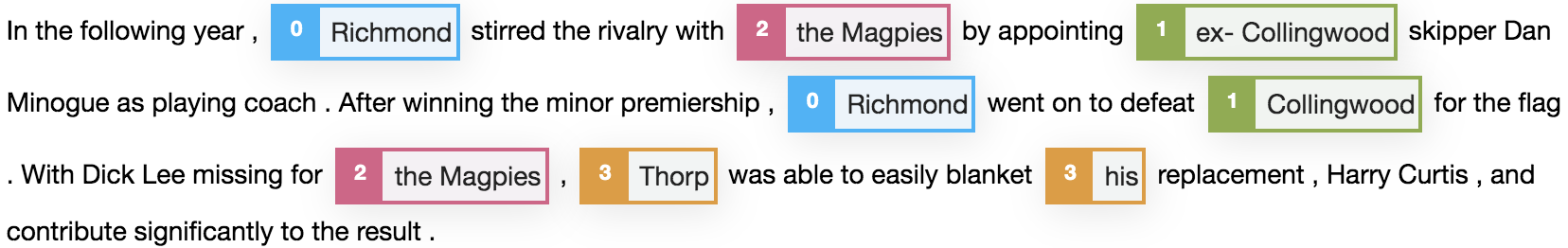}
    }
    \\
\end{tabular}
\caption{\label{table:example3vis} Example 3 - Illustration of ground truth and coreference model predictions. Mentions belonging to a coreference cluster are color coded and indexed.}
\end{table*}

\begin{table*}[t!]
\centering
\begin{tabular*}{2\columnwidth}{@{\extracolsep{\fill}}lccc}
    \hline
         & P(A) & P(B) & P(NEITHER)\\ 
        \hline
        ProBERT & 0.028 &  0.968 & 0.003  \\ 
        GREP & 0.724 & 0.263 & 0.012\\
    \hline
\end{tabular*}
\caption{\label{table:example3probs} Example 3 - A comparison of probabilities assigned by ProBERT and GREP}
\end{table*}

\end{document}